\documentclass{article}

\usepackage[final]{neurips_2021_ml4ps}

\usepackage[utf8]{inputenc} 
\usepackage[T1]{fontenc}   
\usepackage{hyperref}     
\usepackage{url}          
\usepackage{booktabs}     
\usepackage{amsfonts}    
\usepackage{nicefrac}    
\usepackage{microtype} 
\usepackage{xcolor}   
\usepackage{amsmath}
\usepackage{amssymb}
\usepackage{graphicx}
\usepackage{subcaption}

\title{Optimizing High-Dimensional Physics Simulations via Composite Bayesian Optimization}

\author{
Wesley J. Maddox\thanks{Work performed during an internship at Facebook.} \\
New York University \\
\texttt{wjm363@nyu.edu}
\And
Qing Feng \\
Facebook \\
\texttt{qingfeng@fb.com}
\And
Max Balandat \\
Facebook \\
\texttt{balandat@fb.com}
}

\begin{document}

\maketitle

\begin{abstract}
Physical simulation-based optimization is a common task in science and engineering. 
Many such simulations produce image- or tensor-based outputs where the desired objective is a function of those outputs, and optimization is performed over a high-dimensional parameter space. 
We develop a Bayesian optimization method leveraging tensor-based Gaussian process surrogates and trust region Bayesian optimization to effectively model the image outputs and to efficiently optimize these types of simulations, including a radio-frequency tower configuration problem and an optical design problem.
\end{abstract}

\section{Introduction}

Many design problems in the physical sciences require running simulations to evaluate a new design. Examples abound in material science \citep{zhang2020bayesian}, fluid dynamics \citep{anderson1995computational}, and optics~\citep{o2004diffractive}. Typically, these simulations generate high-dimensional outputs, often in the form of image- or other tensor-structured formats. This usually requires substantial computational effort and simulations may take a long time to run. Optimizing designs thus presents a formidable challenge, and using sample-efficient optimization methods is crucial. 

In this work, we solve a challenging design problem, namely optimizing the geometry and gratings of a diffractive optical element.
In our example the design is parameterized by 177 parameters; each simulation takes about one hour to run and produces an $11 \times 16 \times 16$ output representing the projected image. Our goal is to explore the efficient frontier between efficiency (how much light is delivered) and uniformity (how uniform is the output) of the image generated by the device.

One approach to optimizing such a problem is to build a surrogate model of the simulation that is cheap/quick to evaluate, and perform the optimization utilizing this surrogate. 
Bayesian Optimization (BO) is an established method following this paradigm, and has been successfully applied in wide range of settings, including many in the physical sciences \citep{packwood2017bayesian,zhang2020bayesian,PhysRevLett.124.124801}. 
Historically, BO has been restricted to relatively low-dimensional design spaces, a small number of evaluations, and a single scalar outcome. 
Recently, \citet{maddox2021bayesian} developed a scalable approach to \emph{composite} BO with high-dimensional outputs that relies on an efficient sampling scheme for High-Order Gaussian Process (HOGP) models that provide a probabilistic model over tensor-structured outputs. However, their method suffers from the same scaling challenges in terms of the dimension of the \emph{design} space as standard GP models do \citep{eriksson2019scalable,eriksson2020scalable}, and is thus not applicable to our 177-dimensional optics optimization problem.

In this work, we combine the model and sampling scheme from \citet{maddox2021bayesian} with the recent MORBO algorithm for high-dimensional multi-objective Bayesian Optimization from \citet{daulton2021morbo}. We overcome additional memory scalability challenges by employing a mixed-precision compute paradigm and batching computations, enabling improved performance over existing baselines.

\section{Methodology}
\label{sec:CompTurBOHOGP}

\subsection{High-Order Gaussian Process for Image Pixel Prediction} 
\label{subsec:hogp}
To perform composite Bayesian Optimization in the space of large images, we use the high-order Gaussian Process (HOGP) proposed by~\citet{zhe_scalable_2019} to model the images from the optics simulator given a design configuration. This model extends the traditional multi-task Gaussian processes (MTGPs) and can more efficiently handle high-dimensional correlated outputs. 

The HOGP model tensorizes the image outputs $\mathbf y \in \mathbb{R}^{n \times d_1 \times \cdots \times d_k}$, and learns latent features of each tensor element to capture their correlations. It assumes the covariance between any two outputs, $\mathbf y, \mathbf y'$, is the given as the elementwise product of the output indices. The covariance function is $$k([x, i_1, \cdots, i_k], [x', j_1, \cdots, j_k]) = k(x,x') k(v_1, v'_1) \cdots  k(v_k, v'_k),$$ where $i_1, \cdots, i_k$ are the indices for the output tensor, $v_1, \cdots, v_k$ are the latent parameters, and $k(x,x')$ is the kernel over the parameter space. Thus, the task covariance function in the MTGP framework is represented as a chain of Kronecker products so that the GP prior is $\text{vec}(\mathbf y) \sim \mathcal{N}(0, K_{XX} \otimes K_2 \otimes \cdots \otimes K_k).$ See Appendix~\ref{app:kronecker} for more discussions. 

\subsection{Composite Multi-objective Optimization over High-dimensional Search Space}
\label{subsec:morbo}
Given the images predicted from the HOGP model, we can perform composite BO~\citep{astudillo_bayesian_2019} that optimizes composite objectives of the form $\max_{x}g(h(x))$, where $h: \mathbb{R}^d \rightarrow \mathbb{R}^{d_1 \times \cdots \times d_k}$ is the expensive simulation that produces a tensor as output, and $g: \mathbb{R}^{d_1 \times \cdots \times d_k} \rightarrow \mathbb{R}^o$ is the deterministic function to compute goal metric e.g. efficiency of image outputs. 

The optical design problem poses two primary challenges: 1) the design space is high-dimensional with 177 parameters to optimize; 2) the goal is to find the set of optimal tradeoffs between the two competing objectives (efficiency and uniformity) rather than optimizing a single objective. GP-based BO usually works well for problems with search spaces having less than 20 or so parameters, but does not scale well to high-dimensional parameter spaces: as distances grow larger and model uncertainty towards the boundary of the search space increases, BO tends to over-explore. To avoid this issue, \citet{eriksson2019scalable} introduced trust region Bayesian optimization (TRBO) that performs optimization in smaller trust regions that evolve across the search space. More recently, (MORBO) \citet{daulton2021morbo} extended this work to the multi-objective setting. 
While MORBO handles high-dimensional \emph{parameter} spaces well, \citet{daulton2021morbo} only considered problems with few outcomes and non-composite settings. In this work we employ MORBO in conjunction with the improved HOGP model to perform composite BO over high-dimensional \emph{outcome} spaces (images). 

\subsection{Efficient Posterior Sampling for MORBO with HOGP}  
\label{subsec:sampling}
MORBO/TRBO construct trust regions and perform local modeling and optimization. 
Thus, we build a HOGP using observations inside each trust region. Since the HOGP is a sample-efficient model, we can achieve good predictions of image pixels and the aggregated goal metrics (see Figure \ref{fig:diffractive_element_hogp} for an example), and also reduce computational cost of using all the data points.  

Both MORBO and TRBO rely on Thompson sampling for optimizing acquisition functions, which is typically implemented as drawing a large numbers of GP posterior samples evaluated at many discrete candidates, $x_{\textrm{test}}.$ Drawing posterior samples from HOGPs can be computationally expensive and even be intractable for high-dimensional outputs such as images. The time complexity to naively sample over all outputs (tasks) and all new data points is multiplicative in the number of outputs $\mathcal{O}((n^3 + n_{\textrm{test}}^3)\prod_{i=1}^d d_i^3)$~\citep{maddox2021bayesian}. What's more, storing $k$ posterior samples at $n_{\textrm{test}}$ test points requires storage of $k \times n_{\textrm{test}} \times n \times d_1 \times \cdots \times d_k$ tensors, which quickly becomes problematic on a memory-restricted GPU. To make it feasible to combine HOGP with MORBO/TRBO, we leverage the efficient posterior sampling developed by~\citet{maddox2021bayesian} to reduce the time complexity to $\mathcal{O}((n^3 + n_{\textrm{test}}^3) + \sum_{i=1}^d d_i^3)$ (see Appendix~\ref{app:hopg_sampling} for additional technical details) and further propose two remedies to improve memory complexity of posterior sampling.

First, we segment test points into smaller batches and loop batches of $n' \ll n_{\textrm{test}}$ test points, as they affect the size of the Kronecker matrix vector multiplications more than the number posterior samples, $k$.
This results in smaller matrix vector products and thus reduced memory overheads.
While this is not quite Thompson sampling due to the conditional independence between batches, we found that it does not empirically lead to a large degradation in performance. 
However, we hope to improve on this strategy in future work.

Second, we employ a mixed-precision computing paradigm and use half precision arithmetic to compute the Kronecker matrix vector products when evaluating the posterior samples, while using double precision for the numerically demanding matrix root computation of the potentially poorly conditioned data covariance.
Unlike previous work such as \citet{gardner_gpytorch_2018,maddox2021accurate} that found implementation difficulties when moving to lower precision arithmetic, we perform all training in float and double precision arithmetic, and only compute the posterior Kronecker matrix vector products (which are the memory intensive ones) in half precision.
This enables accurate computations when necessary, while preserving much of the 
speedups gained by using lower precision arithmetic \citep{micikevicius2017mixed}.

\section{Results and Discussion}

\subsection{Single Objective Experiments}
We evaluate our method (HOGP+TRBO) that performs composite TRBO with HOGP model on three single objective optimization problems, and compare with four methods: quasi-random search (Random), expected improvement on the metric (qEI), current TRBO (GP+TRBO) and composite BO with HOGP and expected improvement (HOGP+EI). All the results are the mean and 95\% confidence intervals across 20 trials. See Appendix~\ref{app:single_obj} for additional experimental details. 

\begin{figure}[t!]
\begin{subfigure}{0.32\textwidth}
\centering
\includegraphics[width=\linewidth]{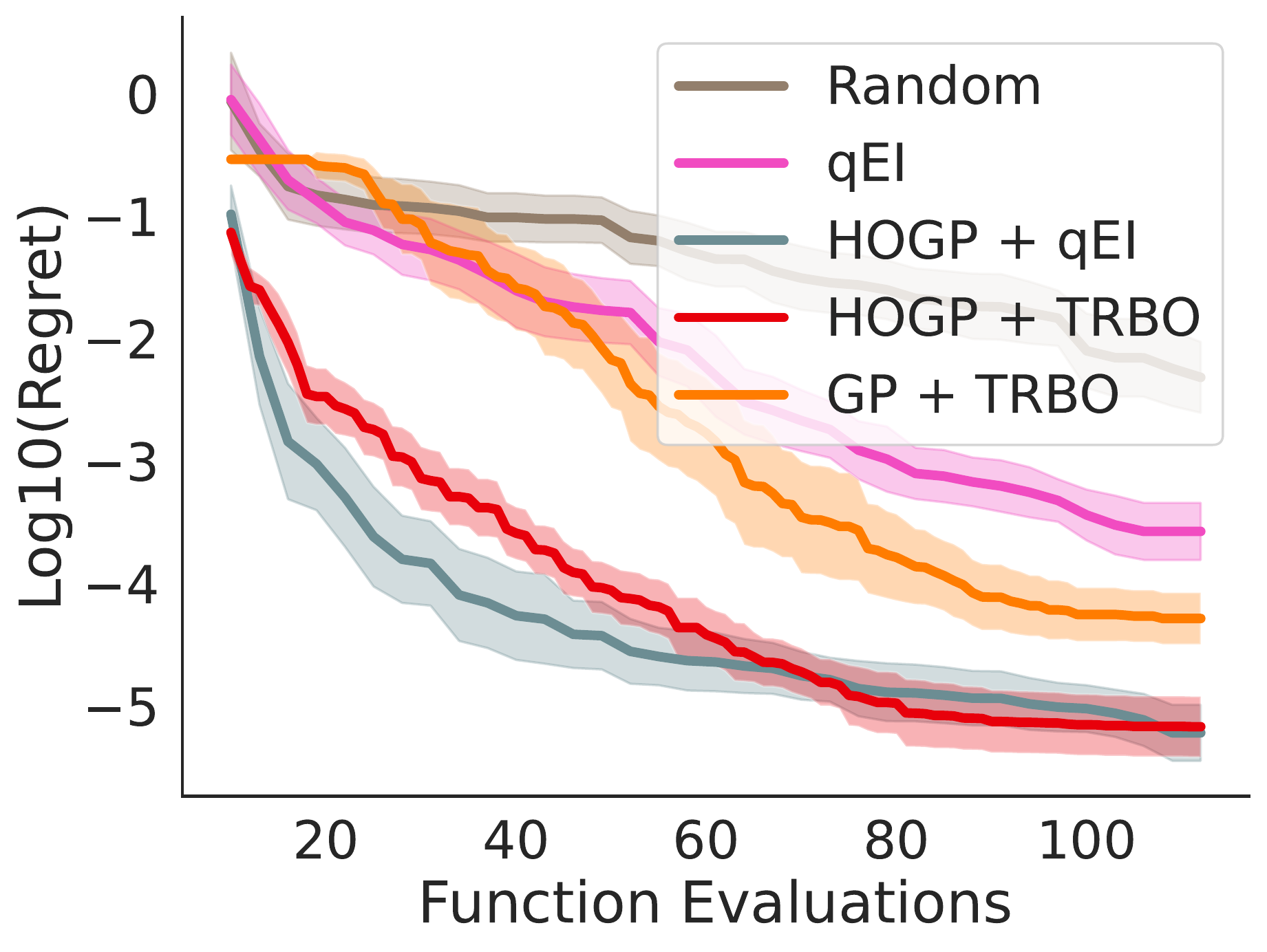}
\caption{Environmental, \newline \centering $t = 3 \times 4, d=4, q=3.$}
\label{fig:environmental_34}
\end{subfigure}
\begin{subfigure}{0.32\textwidth}
\centering
\includegraphics[width=\linewidth]{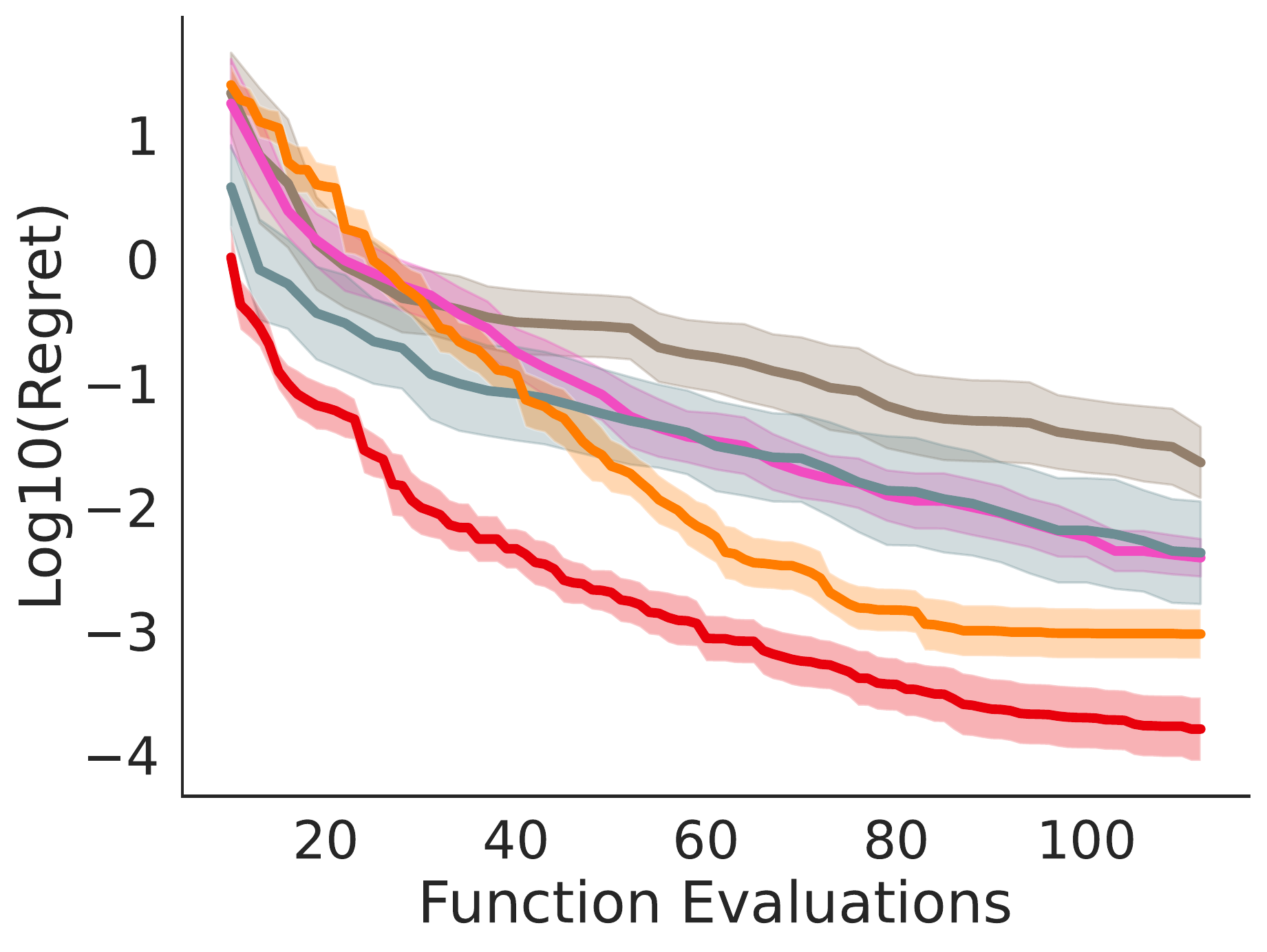}
\caption{Environmental, \newline \centering $t = 5 \times 10, d=4, q=3$.}
\label{fig:environmental_510}
\end{subfigure}
\begin{subfigure}{0.32\textwidth}
\centering
\includegraphics[width=\linewidth]{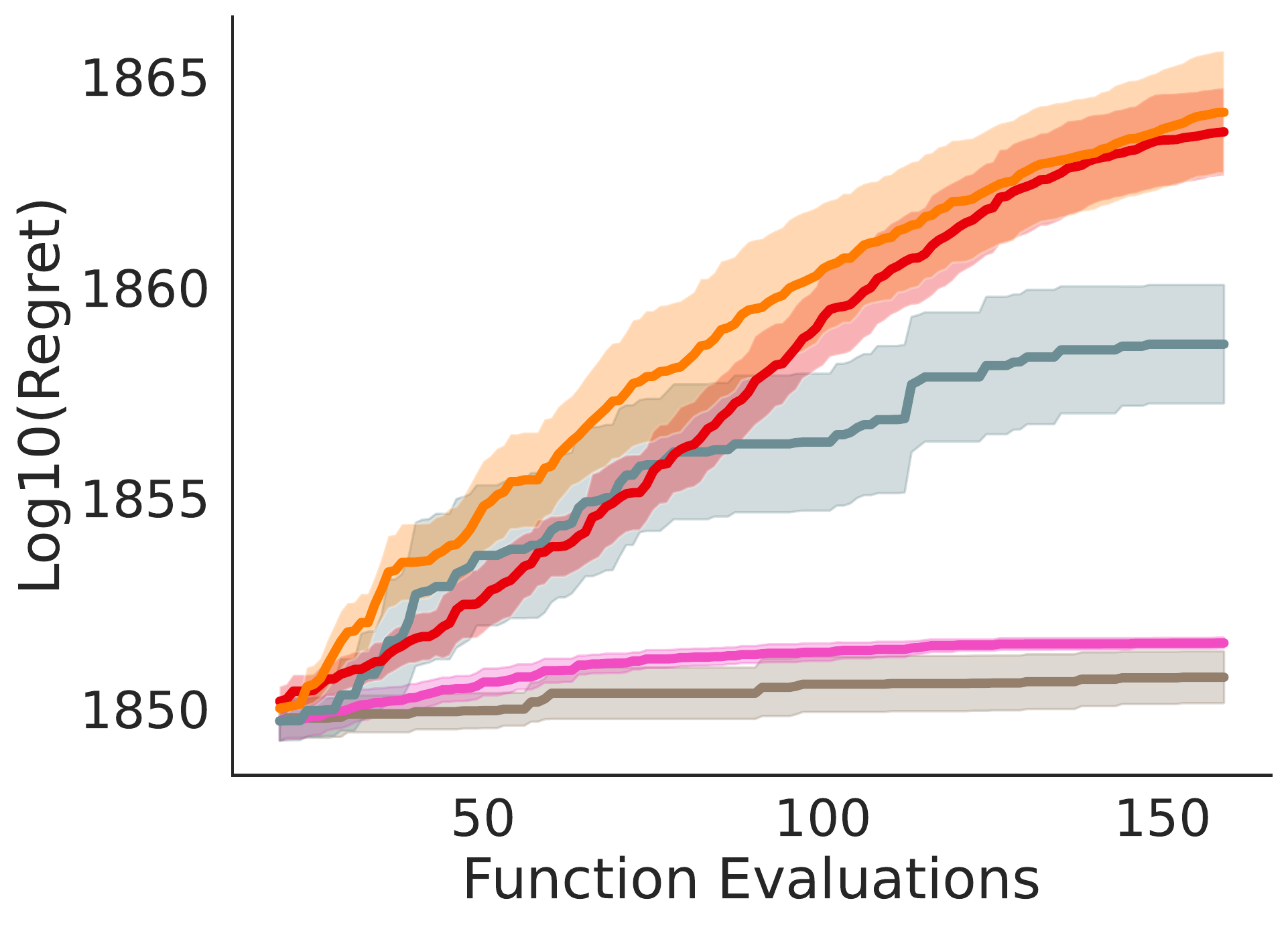}
\caption{Cell-tower coverage, \newline \centering $t = 2 \times 50 \times 50, d=30, q=1.$}
\label{fig:maveric_1}
\end{subfigure}
\caption{Benchmark traces (average across 20 runs with 95\% confidence interval) for single objective problems. HOGP+TRBO achieves the best performance on the environmental problems and performs similarly as TRBO, obtaining the best reward on the cell-tower coverage problem.}
\end{figure}

\paragraph{Environmental Problem}
We evaluate on a spatial problem in which environmental pollutant concentrations are observed on a $3 \times 4$ grid originally defined in \citet{bliznyuk2008bayesian,astudillo_bayesian_2019} and an expanded $5 \times 10$ grid. The goal is to optimize a set of four parameters to achieve the true observed value by minimizing the mean squared error of the output grid to the output grid of the true parameters. As shown in Figure~\ref{fig:environmental_34} and Figure~\ref{fig:environmental_510}, current TRBO achieves lower regret compared with Random and qEI and our method further outperforms TRBO. This demonstrates the efficiency of performing composite Bayesian optimization with HOGP.

\paragraph{Cell-Tower Coverage Problem}
Following~\citet{maddox2021bayesian,dreifuerst2020optimizing}, we optimize the simulated “coverage map"
resulting from the transmission power and down-tilt settings of 15 cell towers (for a total of 30 parameters) based on a scalarized quality metric combining signal power and inference at each location so as to maximize total coverage, while minimizing total interference. To reduce model
complexity, we down-sample the simulator output to $50 \times 50$, initializing the optimization with 20 random configurations. Figure~\ref{fig:maveric_1} shows that our method and current TRBO achieve the best performance.

\subsection{Multi-Objective Design of a Diffractive Optical Element}
We compare our multi-objective optimization approach (HOGP+MORBO) with current MORBO (GP+MORBO) \citep{daulton2021morbo} on the 177-dimensional optics optimization problem.  
The simulations generating the outputs of the elements are computed using a custom physics simulation engine. In order to evaluate the optimization performance at reasonable computational cost, we fit a neural network surrogate model from the input parameterization to the image output based on a large number of simulation runs sampled from the design space. In the benchmarks, we evaluate designs based on this neural network surrogate rather than the real physics simulation engine. See Appendix~\ref{app:optical_design} for the details of optical experimental setup. The goal is to jointly minimize two goal metrics, efficiency and uniformity, used for measuring displayed image quality. We assess the performance based on the maximum achieved hyper-volume. 

\begin{figure}[t!]
\begin{subfigure}{0.49\textwidth}
\centering
\includegraphics[width=0.975\textwidth]{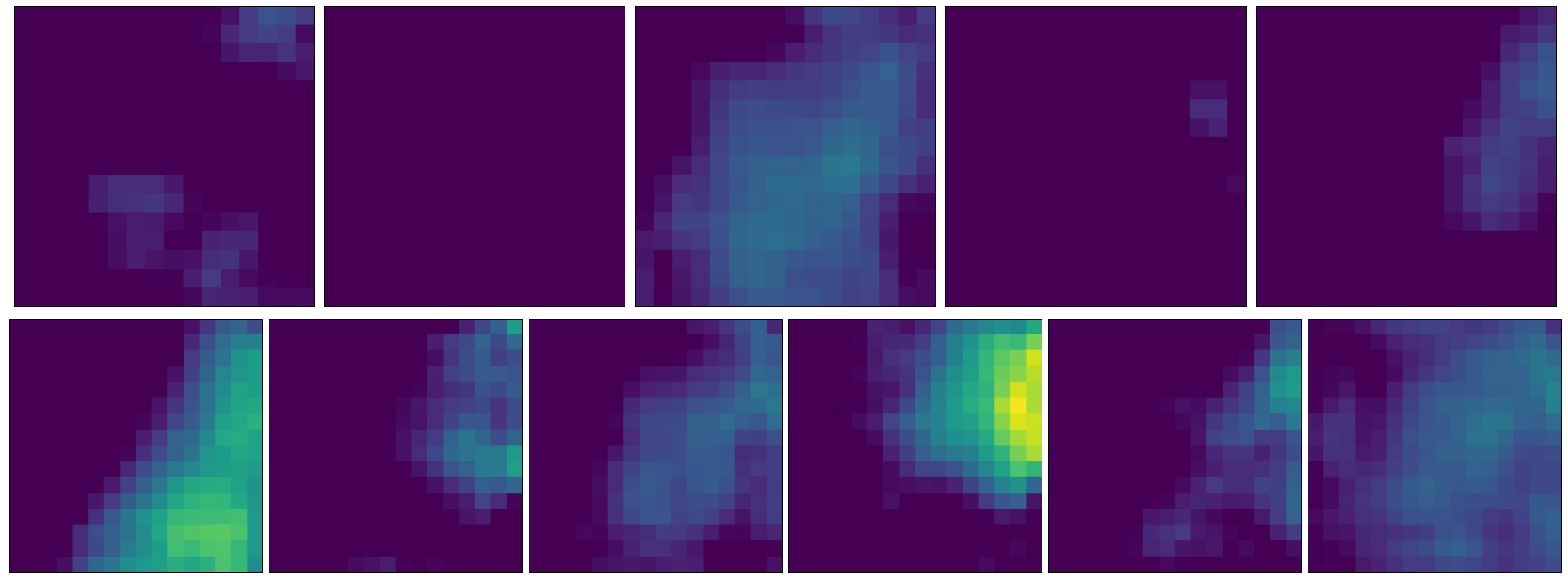}
\caption{Example image outputs from simulations.}
\label{fig:diffractive_element_images}
\end{subfigure}
\begin{subfigure}{0.49\textwidth}
\centering
\includegraphics[width=0.975\textwidth]{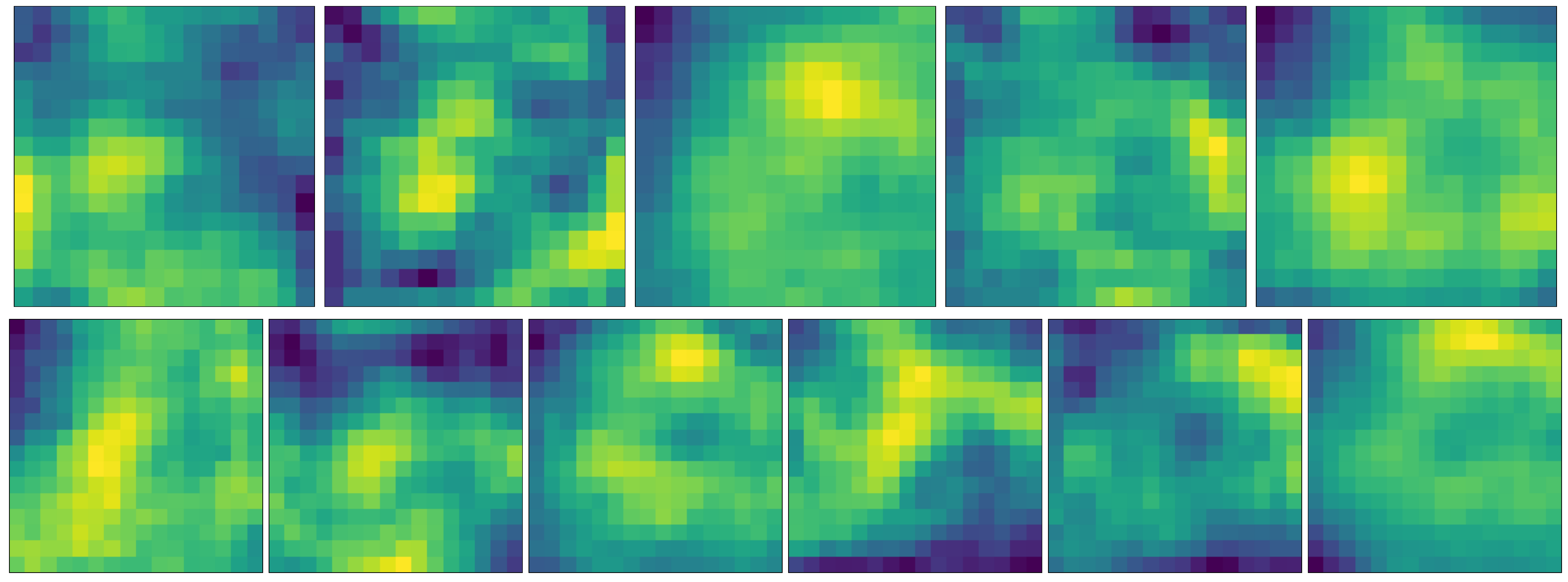}
\caption{Example optimized image outputs from simulations.}
\label{fig:diffractive_element_images_optimized}
\end{subfigure}

\caption{Example output images (un-optimized \textbf{(a)} and optimized \textbf{(b)} using our approach (HOGP+MORBO) on the diffractive optical element problem; pixel values are displayed on the same scale per panel. The ideal output would be uniformly bright images. The optimized channels are significantly brighter and are smoother at same time as would be expected from optimizing both the efficiency and uniformity metrics.}
\end{figure}

\begin{figure}[t!]
\centering
\begin{subfigure}{0.475\textwidth}
\centering
\includegraphics[width=0.975\linewidth]{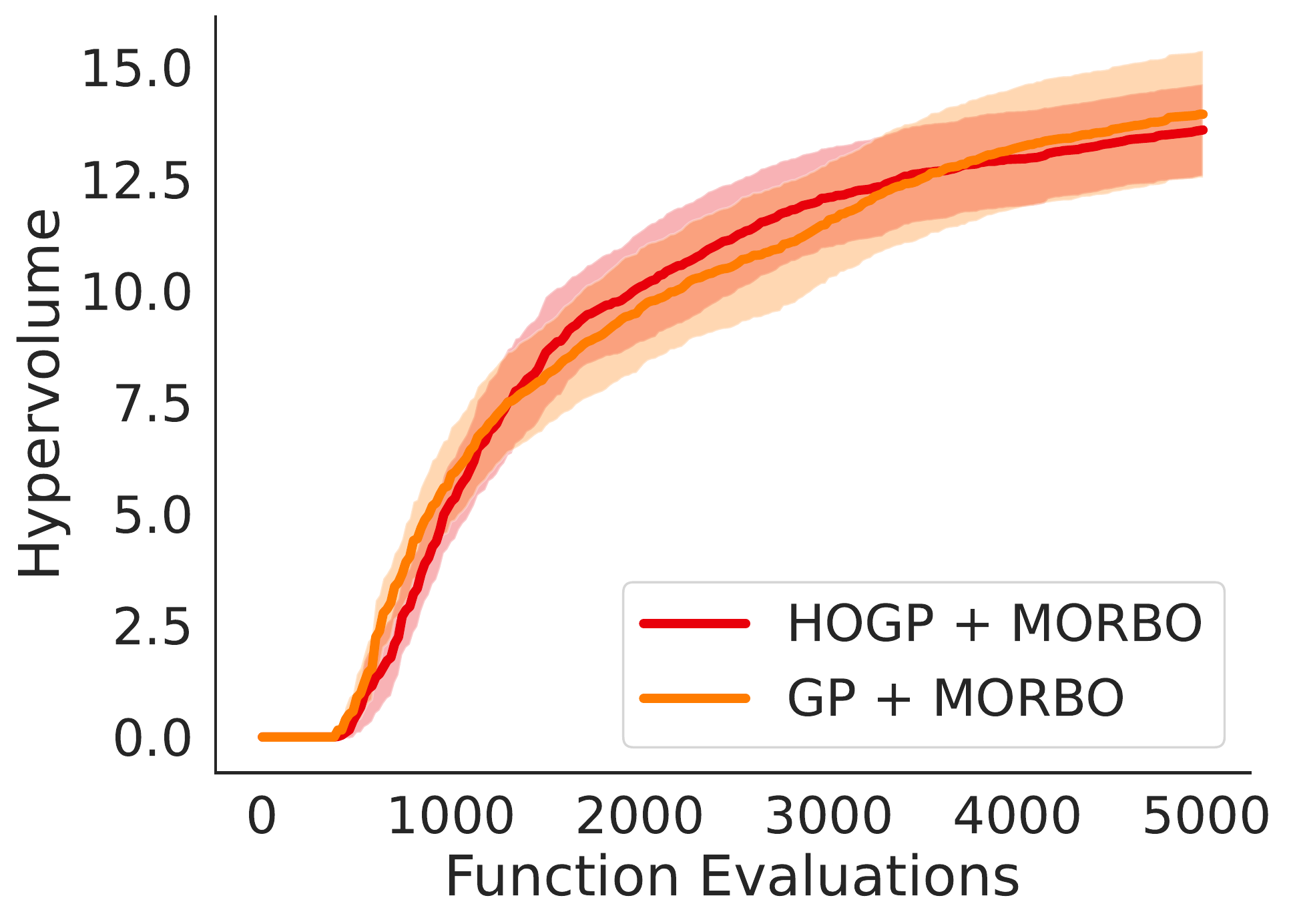}
\caption{}
\label{fig:diffractive_element_noisy}
\end{subfigure}
\begin{subfigure}{0.475\textwidth}
\centering
\includegraphics[width=0.975\linewidth]{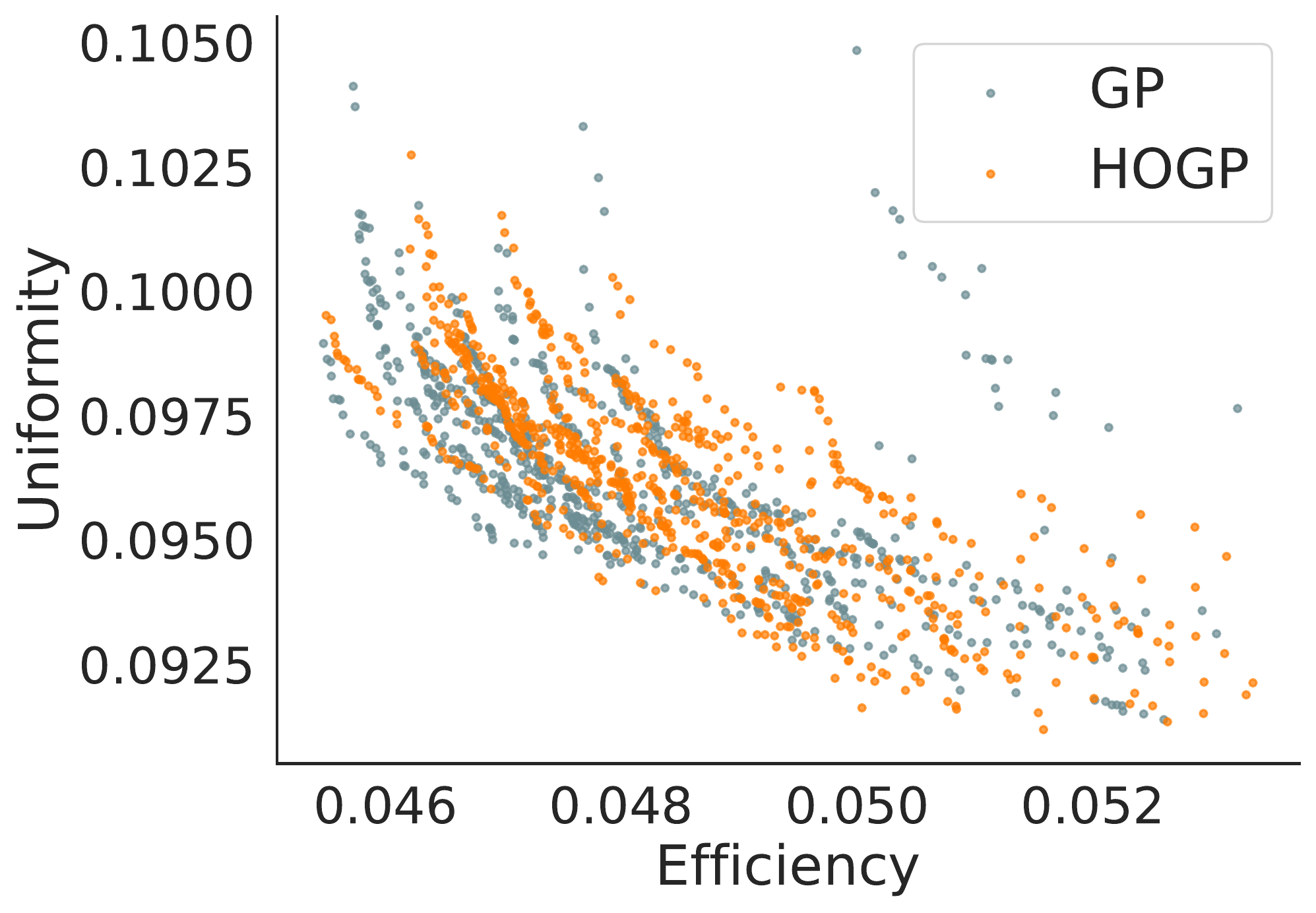}
\caption{}
\label{fig:diffractive_element_pareto}
\end{subfigure}
\caption{Performance on the multi-objective optical design problem. (a) HOGP+MORBO outperforms GP+MORBO in the early iterations and achieves similar maximum hypervolume at the end. (b) Pareto frontiers (metrics are standardized so that lower is better for both metrics) over $20$ trials. HOGP+MORB can push more on optimizing uniformity which is harder to model.
}
\end{figure}

Figure \ref{fig:diffractive_element_images_optimized} demonstrates the substantial improvements on the displayed images through optimizing with HOGP+MORBO. The optimized images are much brighter and are also smoother compared to the un-optimized outputs. Figure \ref{fig:diffractive_element_noisy} shows that our approach tends to outperform current MORBO in the earlier stage of the optimization and reaches to a similar performance at the end. We further visualize the Pareto frontiers (metrics are standardized) across 20 runs in Figure~\ref{fig:diffractive_element_pareto}. It can be seen that the HOGP can explore along the uniformity metric more efficiently than the GP model, but explores less on the efficiency metric. Since the uniformity metric (defined as a ratio between extreme pixel values) is a harder to model than efficiency (which is closely related to the average pixel brightness), this again suggests the sample efficiency of HOGP from modeling the image outputs directly and learning the latent correlation structure across image pixels to transfer information.

\section*{Acknowledgements}
WJM was partially supported by a NSF Graduate Research Fellowship under NSF IIS-1951856. We'd like to thank Dominic Meiser, Ningfeng Huang, Eytan Bakshy, and David Eriksson.

\bibliographystyle{apalike}
\bibliography{references}

\begin{thebibliography}{}

\bibitem[Anderson and Wendt, 1995]{anderson1995computational}
Anderson, J.~D. and Wendt, J. (1995).
\newblock {\em Computational fluid dynamics}, volume 206.
\newblock Springer.

\bibitem[Astudillo and Frazier, 2019]{astudillo_bayesian_2019}
Astudillo, R. and Frazier, P. (2019).
\newblock Bayesian {Optimization} of {Composite} {Functions}.
\newblock In {\em International {Conference} on {Machine} {Learning}}, pages
  354--363. PMLR.
\newblock ISSN: 2640-3498.

\bibitem[Balandat et~al., 2020]{balandat_botorch_2020}
Balandat, M., Karrer, B., Jiang, D., Daulton, S., Letham, B., Wilson, A.~G.,
  and Bakshy, E. (2020).
\newblock {BoTorch}: {A} {Framework} for {Efficient} {Monte}-{Carlo} {Bayesian}
  {Optimization}.
\newblock In {\em Advances in {Neural} {Information} {Processing} {Systems}},
  volume~33.

\bibitem[Bliznyuk et~al., 2008]{bliznyuk2008bayesian}
Bliznyuk, N., Ruppert, D., Shoemaker, C., Regis, R., Wild, S., and Mugunthan,
  P. (2008).
\newblock Bayesian calibration and uncertainty analysis for computationally
  expensive models using optimization and radial basis function approximation.
\newblock {\em Journal of Computational and Graphical Statistics},
  17(2):270--294.

\bibitem[Daulton et~al., 2021]{daulton2021morbo}
Daulton, S., Eriksson, D., and Balandat, M. (2021).
\newblock Multi-objective bayesian optimization over high-dimensional search
  spaces.
\newblock {\em arXiv preprint arXiv:2109.10964}.

\bibitem[Dreifuerst et~al., 2020]{dreifuerst2020optimizing}
Dreifuerst, R.~M., Daulton, S., Qian, Y., Varkey, P., Balandat, M., Kasturia,
  S., Tomar, A., Yazdan, A., Ponnampalam, V., and Heath, R.~W. (2020).
\newblock Optimizing coverage and capacity in cellular networks using machine
  learning.
\newblock {\em arXiv preprint arXiv:2010.13710}.

\bibitem[Duris et~al., 2020]{PhysRevLett.124.124801}
Duris, J., Kennedy, D., Hanuka, A., Shtalenkova, J., Edelen, A., Baxevanis, P.,
  Egger, A., Cope, T., McIntire, M., Ermon, S., and Ratner, D. (2020).
\newblock Bayesian optimization of a free-electron laser.
\newblock {\em Phys. Rev. Lett.}, 124:124801.

\bibitem[Eriksson et~al., 2019]{eriksson2019scalable}
Eriksson, D., Pearce, M., Gardner, J., Turner, R.~D., and Poloczek, M. (2019).
\newblock Scalable global optimization via local bayesian optimization.
\newblock {\em Advances in Neural Information Processing Systems},
  32:5496--5507.

\bibitem[Eriksson and Poloczek, 2021]{eriksson2020scalable}
Eriksson, D. and Poloczek, M. (2021).
\newblock Scalable constrained bayesian optimization.
\newblock In {\em International Conference on Artificial Intelligence and
  Statistics}. PMLR.

\bibitem[Gardner et~al., 2018]{gardner_gpytorch_2018}
Gardner, J., Pleiss, G., Weinberger, K.~Q., Bindel, D., and Wilson, A.~G.
  (2018).
\newblock {GPyTorch}: {Blackbox} {Matrix}-{Matrix} {Gaussian} {Process}
  {Inference} with {GPU} {Acceleration}.
\newblock In {\em Advances in {Neural} {Information} {Processing} {Systems}},
  volume~31, pages 7576--7586.

\bibitem[Goovaerts et~al., 1997]{goovaerts1997geostatistics}
Goovaerts, P. et~al. (1997).
\newblock {\em Geostatistics for natural resources evaluation}.
\newblock Oxford University Press on Demand.

\bibitem[Maddox et~al., 2021a]{maddox2021bayesian}
Maddox, W.~J., Balandat, M., Wilson, A.~G., and Bakshy, E. (2021a).
\newblock Bayesian optimization with high-dimensional outputs.
\newblock {\em arXiv preprint arXiv:2106.12997}.

\bibitem[Maddox et~al., 2021b]{maddox2021accurate}
Maddox, W.~J., Kapoor, S., and Wilson, A.~G. (2021b).
\newblock When are iterative gaussian processes reliably accurate?
\newblock In {\em OPTML Workshop at International Conference on Machine
  Learning (ICML)}.

\bibitem[Micikevicius et~al., 2017]{micikevicius2017mixed}
Micikevicius, P., Narang, S., Alben, J., Diamos, G., Elsen, E., Garcia, D.,
  Ginsburg, B., Houston, M., Kuchaiev, O., Venkatesh, G., et~al. (2017).
\newblock Mixed precision training.
\newblock {\em arXiv preprint arXiv:1710.03740}.

\bibitem[O'Shea et~al., 2004]{o2004diffractive}
O'Shea, D.~C., Suleski, T.~J., Kathman, A.~D., and Prather, D.~W. (2004).
\newblock {\em Diffractive optics: design, fabrication, and test}, volume~62.
\newblock SPIE press.

\bibitem[Packwood, 2017]{packwood2017bayesian}
Packwood, D. (2017).
\newblock {\em Bayesian Optimization for Materials Science}.
\newblock Springer.

\bibitem[Rasmussen and Williams, 2008]{rasmussen_gaussian_2008}
Rasmussen, C.~E. and Williams, C. K.~I. (2008).
\newblock {\em Gaussian processes for machine learning}.
\newblock Adaptive computation and machine learning. MIT Press, Cambridge,
  Mass., 3. print edition.

\bibitem[Zhang et~al., 2020]{zhang2020bayesian}
Zhang, Y., Apley, D.~W., and Chen, W. (2020).
\newblock Bayesian optimization for materials design with mixed quantitative
  and qualitative variables.
\newblock {\em Scientific reports}, 10(1):1--13.

\bibitem[Zhe et~al., 2019]{zhe_scalable_2019}
Zhe, S., Xing, W., and Kirby, R.~M. (2019).
\newblock Scalable {High}-{Order} {Gaussian} {Process} {Regression}.
\newblock In {\em The 22nd {International} {Conference} on {Artificial}
  {Intelligence} and {Statistics}}, pages 2611--2620. PMLR.
\newblock ISSN: 2640-3498.

\end{thebibliography}

\appendix

\section{Broader Impact Statement}
The method introduced in this paper provides a sample-efficient solution to the challenging design problems in the physical sciences. By overcoming the scalability blockers in leveraging tensor-based Gaussian Process model and trust region Bayesian Optimization, we unlock the possibilities of conducting optimizations in high-dimensional parameter spaces and output spaces. 
We do not expect any negative social impacts from our work.

\section{Methodological Details}

\subsection{Kronecker Matrix Vector Products}
\label{app:kronecker}
The key aspect of efficiency in the HOGP comes from \emph{Kronecker matrix vector products}. Such structure allows it to capture complex output correlations and scale to high-dimensional outputs with no sparse approximation. Besides, kronecker matrix vector multiplies (MVMs) can be efficiencly computed from:
\begin{align*}
	z = (K_1 \otimes K_2)\text{vec}(A) = \text{vec}(K_2 A K_1^\top);
\end{align*}
if $K_1 \in \mathbb{R}^{n_1 \times n_1}$ and $K_2 \in \mathbb{R}^{n_2 \times n_2}.$ As a result, computing $z$ costs $\mathcal{O}(n_1^2 + n_2^2 + n_1 n_2 (n_1 + n_2))$ time.
Note that we can recursively compute this structure across several matrix vector products.
Implementation wise, this involves reshaping the vector $\text{vec} A$ to be a matrix and then computing a matrix matrix product, for example $K_2 A$.

\subsection{Efficient Posterior Sampling with the HOGP}
\label{app:hopg_sampling}
 We utilize the efficient posterior sampling mechanism for the HOGP model \citet{maddox2021bayesian} proposes an efficient posterior sampling mechanism for general MTGPs and extends it to HOGPs as a special case. The time complexity of this method is additive in the combination of tasks and data points, rather than multiplicative, which allows us to perform time-efficient composite Bayesian Optimization on the image outputs.  

This method uses Matheron's rule for sampling conditional Gaussian distributions~\citep{goovaerts1997geostatistics}. For HOGP, $f(x_{\text{test}}) | Y = y$ generated by Matheron's rule can be represented as 
\begin{align}
  	\bar{f} = f + \left(K_{x_{\text{test}} X}\otimes_{i=2}^d K_{i}\right)\left((K_{XX} \otimes_{i=2}^d K_{i}) + \sigma^2 I\right)^{-1}(y - Y - \epsilon),  
\end{align}
where $f \sim \mathcal{N}(0, K_{(x_{\text{test}}, X), (x_{\text{test}}, X)} \otimes_{i=2}^d K_i),$ that is drawn from the joint prior distribution that all kernel matrices are Kronecker structured, and $\epsilon \sim \mathcal{N}(0, \sigma^2 I).$ Although the size of $\bar{f}$ is still $\sum_{i=1}^k d_i$ and naively decomposing the posterior covariance matrix of the GP to produce posterior samples would cost $\mathcal{O}((\sum_{i=1}^k d_i)^3)$ time, the Matheron's rule approach instead costs $\mathcal{O}(\sum_{i=1}^k d_i^3 + \sum_{i=1}^k d_i)$ time to draw a single sample. 

\section{Experimental Details}
\subsection{Single Objective Experimental Details}
\label{app:single_obj}
In the benchmarks, the HOGP model and TRBO used reference implementations from their authors with default settings. For qEI and GP+TRBO, we used a standard ARD Matern 5/2 kernel~\citep{rasmussen_gaussian_2008} to model the aggregated goal metrics. For the methods using expected improvement, we used the \textit{qExpectedImprovement} acquisition implemented in BoTorch~\citep{balandat_botorch_2020}.

For the environmental problem, we followed the implementations of \citet{balandat_botorch_2020}, \citet{astudillo_bayesian_2019}, and used 8 random restarts, 256 MC samples, and 512 base samples, a batch limit of 4, and an initialization batch limit of 8. 

For the radio frequency coverage problem, we followed the evaluation setup in \citet{maddox2021bayesian}. We initialized with 20 points, down-sampled the two
$241 \times 241$ outputs to $50 \times 50$ for simplicity, ran the experiments over 20 random seeds and for 150 steps. We used 32 MC samples, 64 raw samples with a batch limit of 4 and an initialization batch limit of 16.

\subsection{Diffractive Optical Element Design}
\label{app:optical_design}
The surrogate model is a densenet style architecture and maps the $177$ dimensional input parameters into the $11$ output images, each of which is of size $16 \times 16.$
Example images that are optimized are shown in Figure \ref{fig:diffractive_element_images}.
The goal of these design efforts is to jointly optimize the efficiency and uniformity of the images. 
Efficiency is computed as a weighted mean across an images, while uniformity is computed as a ratio of the $99\%$ percentile pixel in the image to the $1\%$ percentile pixel in the image.
To combine the metrics across images, we consider the log sum exp of all of the images as a form of soft maximum.\footnote{Please note that all scales of metrics in plots are normalized to be approximately $[0, 1].$}
As the real simulation itself is noisy, we use a scale Binomial noise term to inject noise into the image after being outputted from the surrogate NN mmodel: each image pixel is drawn from a distribution following $y \sim \text{Binomial}(N, p * 100) / (100 * N),$ where $N = 5000.$ 
This type of noise model matches the underlying physical structure of the simulator.

\paragraph{Experimental Details} We used the MORBO implementation as referenced in \citet{daulton2021morbo}. Each trial was initialized with $400$ Sobol points and optimized the hypervolume improvement with batch sizes of $50$ over $5000$ function evaluations. 
The results shown in Figure \ref{fig:diffractive_element_noisy} are the mean and 95\% confidence intervals (2 standard errors) of the achieved maximum hyper-volume across 20 trials.
 
\paragraph{Benchmark Result Analysis} 

\begin{figure}[h!]
\centering
\begin{subfigure}{\textwidth}
\centering
\includegraphics[width=0.95\linewidth]{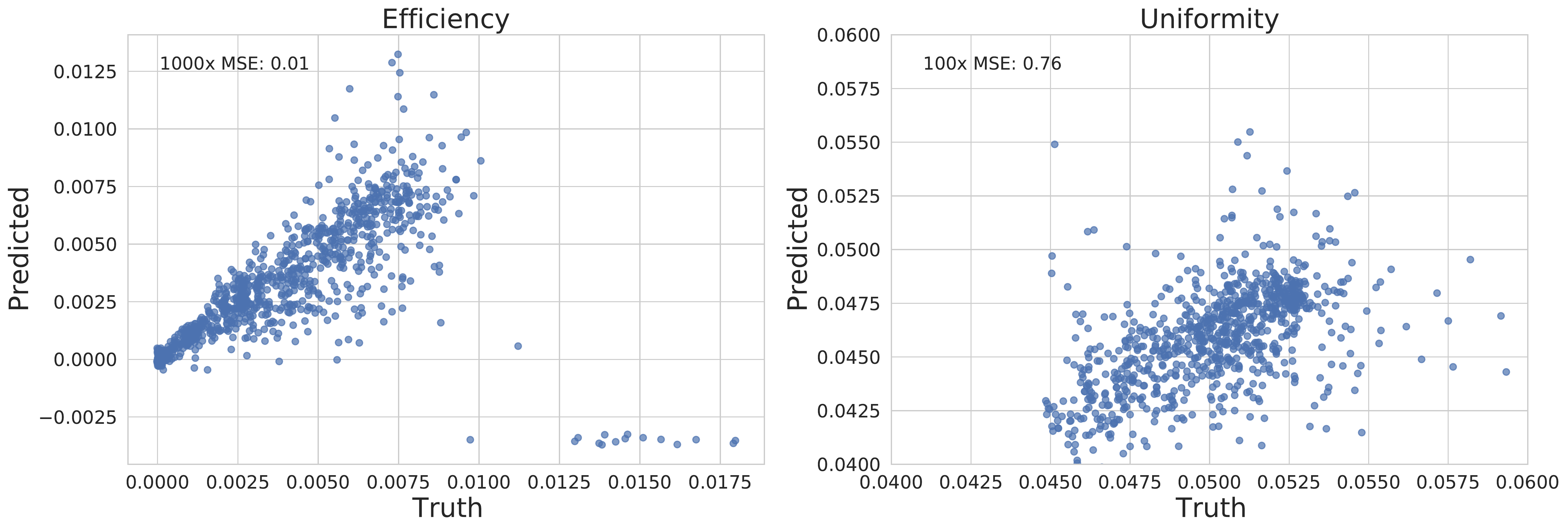}
\caption{HOGP model fits on metrics after $5000$ trials.}
\label{fig:diffractive_element_hogp}
\end{subfigure}
\begin{subfigure}{\textwidth}
\centering
\includegraphics[width=0.95\linewidth]{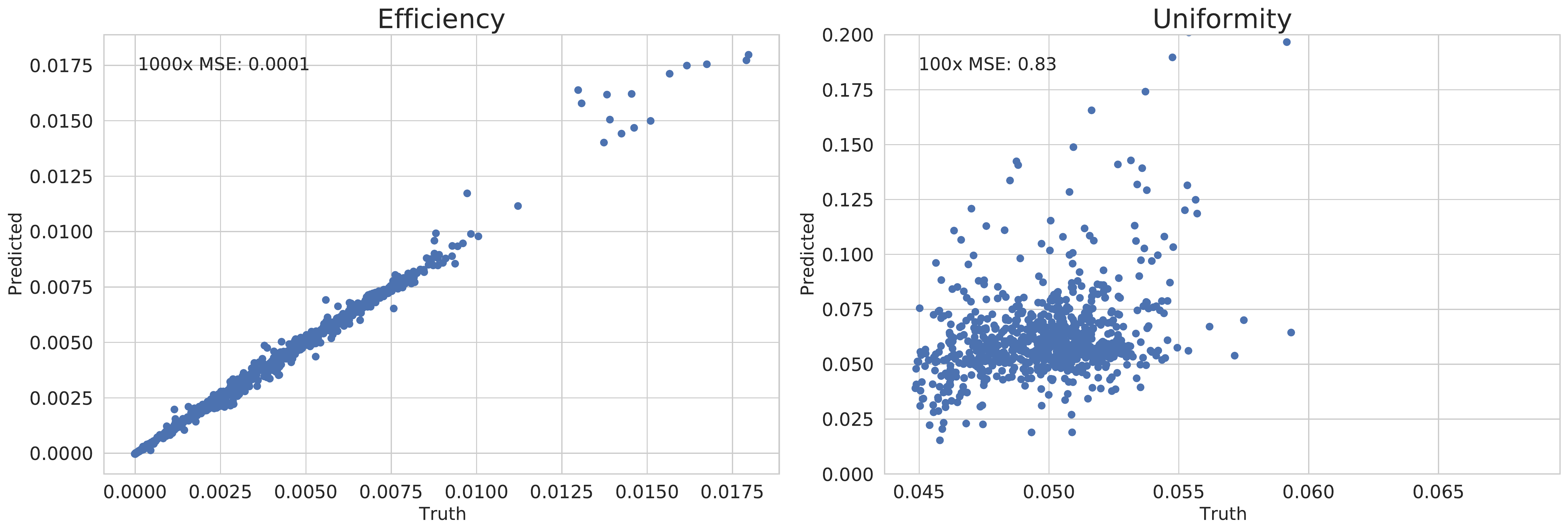}
\caption{GP model fits on metrics after $5000$ trials.}
\label{fig:diffractive_element_gp}
\end{subfigure}
\caption{Global surrogate model fits on the optics problem at the end of an optimization run (test / train sets are the same for both models). The GP surrogate is better at modelling efficiency, but is worse at modelling uniformity. The uniformity metric requires information across all pixels to be modelled accurately, so the HOGP is a better surrogate as it models each pixel.}
\end{figure}

We evaluate the out-of-sample prediction accuracy of standard GP and HOGP model shown in Figures \ref{fig:diffractive_element_hogp} and \ref{fig:diffractive_element_gp}. The plots compare the prediction of two goal metrics on a holdout set of data from an optimization run. We see that while both models are reasonably accurate at predicting both metrics, the GP is more accurate at predicting the efficiency metric, explaining why it was able to explore across the Pareto frontier better on that metric. By comparison, the HOGP is able to use the information of all pixels to better predict the uniformity metric, which is tougher to model and more bi-modal --- we truncated the test set to remove the outliers.

\end{document}